\documentclass[11pt,a4paper]{article}
\usepackage[hyperref]{acl2019}
\usepackage{times}
\usepackage{latexsym}
\usepackage{graphicx}
\usepackage{tikz}
\usepackage{booktabs}
\usepackage{amsmath}
\usepackage{amssymb}
\usepackage{comment}

\usepackage{subcaption}

\usepackage{url}

\aclfinalcopy 

\newcommand\BibTeX{B\textsc{ib}\TeX}

\title{
Dissecting 
\textit{Content} and \textit{Context} in Argumentative Relation Analysis
}

\author{Juri Opitz {\normalfont and} Anette Frank \\
  Research Training Group AIPHES, \\
  Leibniz ScienceCampus ``Empirical Linguistics and Computational Language Modeling''\\
  Department for Computational Linguistics \\
  69120 Heidelberg \\
 {\tt $\lbrace$opitz,frank$\rbrace$@cl.uni-heidelberg.de} }
 
\date{}

\begin{document}
\maketitle
\begin{abstract}
  When assessing relations between argumentative units (e.g., \textit{support} or \textit{attack}), computational systems often exploit disclosing indicators or markers that are not part of elementary argumentative units (EAUs) themselves,
  but are gained from their
  context (position in paragraph, preceding tokens, etc.). We show that this dependency is much stronger than previously assumed. 
  In fact, we show that 
  by completely masking 
  the EAU text spans 
  and only feeding information from 
  their context, 
  a competitive system may function even \textit{better}. 
We argue that an argument analysis system that relies more on discourse
context than the argument's content is unsafe, since it can easily be tricked.
  To alleviate this issue, we separate
  argumentative units from their \textit{context} such that the system is
  forced to 
  model and rely on an EAU's \textit{content}.
  We show that the resulting 
  classification system is 
  more robust, and argue that such models are better suited for
  predicting argumentative relations across documents.

\end{abstract}

\section{Introduction}

In recent years we have witnessed a 
great 
surge in activity in the area of computational argument analysis (e.g.\ \citet{peldszus2013argument,stab2014identifying,rasooli-tetrault-2015,stab2018cross}), 
and the
emergence of dedicated venues such as the ACL Argument Mining workshop series 
starting 
in 2014 
 \citep{W14-21:2014}. 

Argumentative relation classification is a sub-task of argument analysis that aims to determine
relations 
between 
argumentative units A and B, for example, A \textit{supports} B; A \textit{attacks} B. 
Consider the following 
argumentative units \textbf{(1)} and \textbf{(2)}, 
given the topic \textbf{(0)} \textit{``Marijuana should be legalized''}:

\begin{description}
    \item[(1)] \textit{Legalizing marijuana can increase use by teens, with harmful results.}
    \item[(2)] \textit{Legalization allows the government to set age restrictions on buyers.}
    \label{ex:1}
\end{description}

\begin{figure}
    \centering
    \includegraphics[scale=0.7]{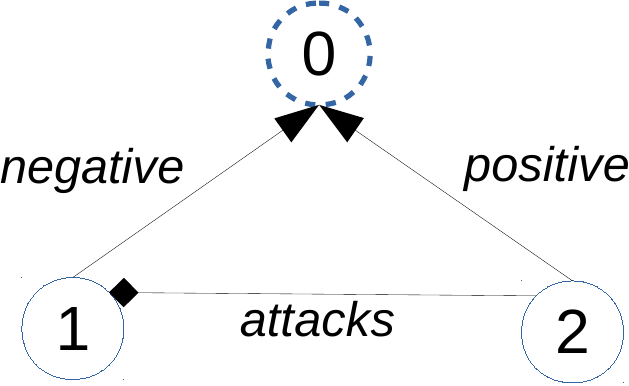}
    \caption{A graph representation of a
    topic (node w/ dashed line), two argumentative premise units (nodes w/ solid line), premise-topic relations (positive or negative) and premise-premise relations
    (here: \textit{attacks}).
    }
    \label{fig:graph4example}
\end{figure}

This example is modeled in Figure \ref{fig:graph4example}.

It is clear that \textbf{(1)} has a negative stance towards the topic and \textbf{(2)} has a positive stance towards the topic. Moreover,  we can say that \textbf{(2)} \textit{attacks} \textbf{(1)}. In discourse, such a relation is often made explicit through discourse markers: \textit{\textbf{(1)}.\ However, \textbf{(2)}; On the one hand \textbf{(1)}, on the other \textbf{(2)}; \textbf{(1)}, although \textbf{(2)};  Admittedly, \textbf{(2)}}; etc. In the absence of such markers we must determine 
this relation by assessing the semantics of the individual argumentative units, including (often implicit)
world knowledge about how they are related to each other.\footnote{In case of \textbf{(1)} and \textbf{(2)}: By setting age restrictions on legalization of Marijuana, increased use by teens can be (expected to be) prevented, thus we can infer that \textbf{(2)} attacks \textbf{(1)}.}

In this work, we show that argumentative relation classifiers -- when provided with textual \textit{context} surrounding an argumentative unit's span -- are very prone to neglect the actual textual \textit{content} of the EAU span. Instead they heavily rely  on \textit{contextual markers}, such as conjunctions or adverbials, as a basis for prediction.
We argue that a system's capacity of predicting the correct relation based on the argumentative units' \textit{content} is important in many circumstances, e.g., when 
an argumentative debate crosses document boundaries.
For example, the \textit{prohibition of marijuana} debate extends across populations and countries -- argumentative units for this debate can
be recovered from thousands of documents scattered across the world wide web. 
As a consequence, argumentative relation classification systems should not be (immensely) dependent on contextual clues -- in the discussed cross-document setting these clues may even be misleading for
such a system,  since source and target arguments can be embedded in 
different
textual contexts (e.g., when \textbf{(1)} and \textbf{(2)} stem from different documents it is easy to imagine a textual context where \textbf{(2)} is not 
introduced
by \textit{however} but
instead
by an `inverse' form such as e.g.\ \textit{moreover}).

\paragraph{Contributions} In Section \S \ref{sec:argmodels} we describe argumentative relation classification systems and their features. Then, to assess the systems' dependency on context, we propose a three-way feature grouping: (i) features which access only the EAU span; (ii) features which access only the context of an EAU; (iii) features which access both EAU span and its context. 
Our experimental results (\S \ref{sec:experiments}) indicate that systems, when given the option, 
tend to
focus on the \textit{context} of an EAU, while neglecting its \textit{content}. 
On the one hand, this leads to strong performance when EAUs appear sequentially in a rhetorically well structured argumentative monologue. Yet, on the other hand, we show that such systems can easily be fooled, e.g., when EAUs are extracted from different documents. 

\section{Related Work}

It is well-known that
the rhetorical and argumentative structure of texts bear great similarities. 
For example, \citet{Azar1999,green2010representation,peldszus2013argument} observe that elementary discourse units (EDUs) in RST \citep{mann1987rhetorical} share great similarity with elementary argumentative units (EAUs) in argumentation analysis.\footnote{Throughout this work we often drop ``elementary'' and use the phrases \textit{EAU and (elementary) argumentative unit} and  \textit{argumentative component} interchangeably.} \citet{wachsmuth2018argumentation} experiment with a modified
version of the Microtext corpus \citep{PeldszusStede-ECA:16}, which is an extensively annotated albeit small corpus. Similar to us, they separate
argumentative units from discursive contextual markers. %
While \citet{wachsmuth2018argumentation}
conduct a \textit{human} evaluation to investigate the separation of \textit{Logos} and \textit{Pathos} aspects of arguments, 
our work investigates
how (de-)contextualization of argumentative units affects \textit{automatic} argumentative relation classification models. 

\paragraph{Notions of context} 
Various notions of \textit{context} are being used in the area of argumentation mining.
For example, \citet{lippi2016argumentation} develop a \textit{context-independent} claim detection system, where by \textit{context-independent} they refer to a system which is not
tailored 
to a specific topic (analogously, \citet{levy-etal-2014-context} aim at \textit{context-dependency}). 
Another notion of \textit{context} concerns the graph context in which relations and EAUs are embedded \cite{kuri-2018}. On the other hand, we adopt a more \textit{textual} notion of context, that is we take a given EAU span as content and text which is not in the EAU span as context. 
This goes in the same direction as \citet{stab2014identifying,DBLP:journals/corr/StabG16,persing-ng-2016-end} and \citet{aker-etal-2017-works} who incorporate features derived from EAU-surrounding text in their classification systems. However, they do not clearly separate between a word indicator feature extracted from within 
(or outside) the EAU span.
For example, when computing features for an EAU, they also take into account EAU-preceding tokens. The preceding tokens, often contain shallow discourse markers which highlight the relationship between two EAUs (e.g., \textit{because}, \textit{however}, etc.).

To the best of our knowledge,
prior work has not yet thoroughly investigated the impact of features extracted
from the EAU vs.\ features extracted from the EAU-embedding context. Our
work
fills this gap 
and shows that the  impact of 
contextual clues from the EAU \textit{context} on classifier performance can be much greater than the impact of features extracted from the
EAU \textit{content}.

\paragraph{Context matters} \citet{nguyen2016context,nguyen2018context} extract additional features from the text between source and target EAUs
 (on the StudentEssay-v01 data \cite{stab-gurevych-2014-annotating}) which results in enhanced predictive performance.
 However, having seen the clear advantages 
 of incorporating context (performance-wise), we find that the downsides of incorporating context remain untold. In this work, we demonstrate that systems which are offered EAU \textit{context} may be  prone to neglect the EAU \textit{content}, an issue that can have undesired effects.

\paragraph{Argumentative relation classification} Argumentative relation classification \citep{mochales2011argumentation} is the task 
for which we aim to examine the context-content relationship. It is concerned with predicting and analyzing relations between argumentative units such as, for example, \textit{support} or \textit{attack}. Besides works discussed above \citep{nguyen2016context,stab2014identifying,DBLP:journals/corr/StabG16}, this task has also been addressed by \citet{cocarascu2017identifying} who develop a neural model to label the edge between two EAUs with $\{attack,support,\varnothing \}$. 
The task has also been approached 
by taking
global graph context into account. E.g., \citet{hou2017argument} jointly model 
argument relation classification and stance classification 
in the DebatePedia\footnote{\url{http://debatepedia.idebate.org/}} corpus using Markov logic networks \cite{richardson2006markov}.  \citet{peldszus2015joint} 
experiment with Microtexts and show that it can be beneficial to model argumentative relations jointly in a network with a minimum spanning tree decoding algorithm. Our work focuses 
on 
local relation prediction and labeling using
the well-established StudentEssay-v02 data \citep{DBLP:journals/corr/StabG16}\footnote{\url{https://tinyurl.com/y269fq3k}} with 402 argumentative essays and thousands of annotated relations between EAUs.

\section{Argumentative Relation Prediction: Models and Features}
\label{sec:argmodels}
In this section, we describe different formulations of the argumentative relation classification task and describe features used by our replicated model. In order to test our hypotheses, we propose to group all features into three distinct types. 

\paragraph{Three feature types: content-based; content-ignorant; full access} %
We categorize features of \citet{DBLP:journals/corr/StabG16} 
into three 
types: (i) features derived from the \textit{context} of the argumentative unit
(e.g., leading and trailing tokens surrounding the EAU span), (ii) features derived from the argumentative unit's  \textit{content} (i.e., the EAU span), and (iii) a joint feature set consisting 
of the union of features from (i) and (ii). However, 
in (iii) we additionally
include features that capture discourse structures that overlap the boundaries between an EAU and its surroundings.

\paragraph{Notations} Henceforth we 
denote 
models that
only make
use of features of type (i), ignoring anything inside the EAU, as \textbf{content-ignorant ($\mathcal{CI}$)}, and models
that are
given only features covering the EAU span 
as \textbf{content-based ($\mathcal{CB}$)}. 
A
model that
combines both is denoted by \textbf{full-access ($\mathcal{FA}$)}. 
We distinguish these different model types
with
a type-variable 
$\mathcal{T} \in \{\mathcal{CI,CB,FA\}}$.

\subsection{Models}

Now, we introduce 
a classification of three different prediction models used in the argumentative relation prediction literature. We will inspect all of them and show that all can suffer from severe issues when focusing (too much) on the context. 

The model $h$ adopts a discourse parsing view on argumentative relation prediction and predicts one outgoing edge for an argumentative unit (\textbf{one-outgoing edge}). Model $f$ assumes a connected graph with argumentative units and is tasked with predicting edge labels for unit tuples (\textbf{labeling relations in a graph}). Finally, a model $g$ is given two (possibly) unrelated argumentative units and is tasked with predicting connections as well as edge labels (\textbf{joint edge prediction and labeling}).  
\paragraph{One-outgoing edge}
\citet{DBLP:journals/corr/StabG16} divide the task into relation prediction $l$ and relation class assignment $h$:
\begin{equation}
\label{eq:0}
l^\mathcal{T}:A\times A\rightarrow \{linked,\varnothing \}
\end{equation}

\vspace*{-3ex}
\begin{equation}
\label{eq:1}
    h^\mathcal{T}: A\rightarrow \{attack,support \},
\end{equation}

which the authors describe as \textit{argumentative relation identification} ($l$) and \textit{stance detection} ($h$). 
%
In their experiments, $\mathcal{T}=FA$, i.e., 
no distinction is made between 
features 
that access
only the argument content (EAU span)
or only the EAU's embedding context,
and some features also consider both (e.g., discourse features). 
This model adopts a
 parsing view 
on argumentative relation classification: every unit is allowed to have only one type of outgoing relation (this
follows trivially from
the fact that $h$ has only one input). 
Applying such a model to argumentative attack and support relations might impose
unrealistic constraints on the 
resulting argumentation graph: A given premise might in fact attack or support several other premises.\footnote{E.g., \textit{this decision will improve the living situation for children. It may also support elderly people with low income.}}
The approach may suffice
for the 
case of student argumentative essays, where EAUs are well-framed in a discourse structure, but seems overly restrictive 
for many other scenarios.

\paragraph{Labeling relations in a graph}
Another
way of framing the task,
is to learn a function 
\begin{equation}
    f^\mathcal{T}:A\times A\rightarrow \{support,attack\},
    \label{eq:2}
\end{equation}

Here, an argumentative unit is allowed to 
be in a
attack or support relation to multiple other EAUs. 
Yet, both $h$ and $f$ assume 
that inputs are already linked and only
the class of the link is unknown. 

\paragraph{Joint edge prediction and labeling}
Thus, we might 
also 
model the task in a three-class classification setting to learn a more general function that performs relation
prediction and classification jointly (see also, e.g., \citet{lippi2016argumentation}):
\begin{equation}
\label{eq:3}
g^\mathcal{T}:A\times A\rightarrow \{support,attack,\varnothing \}.
\end{equation}

The model described by Eq.\  \ref{eq:3}  
is 
the most
general one: not only does it 
assume a graph view on argumentative units and their relations (as does Eq.\
\ref{eq:2}); in model formulation (Eq.\ \ref{eq:3}), 
an argumentative unit can have no or multiple support or attack relations. 
It naturally allows for cases where an
argumentative unit $a$ (supports $b$ $|$ attacks $c$ $|$ \textit{is-unrelated-to} $d$). Given a set of EAUs mined from different documents,
this model enables us to construct 
a full-fledged argumentation graph.

\subsection{Feature implementation}

Our feature implementation follows
the feature descriptions for \textit{Stance recognition} and \textit{link identification} in \citet{DBLP:journals/corr/StabG16}. These features and variations of them have been used 
successfully 
in several successive works  (cf.\ \citet{stab2014identifying,nguyen2016context,aker-etal-2017-works}). 

For any model the features are indexed by $I=\{1,...,N\}$. 
We create a function $\Phi: I \rightarrow \mathcal{T}$ which maps from feature indices to feature types. In other words, $\Phi$ tells us, for any given feature,
whether it is content-based ($\mathcal{CB}$), content-ignorant ($\mathcal{CI}$) or full access ($\mathcal{FA}$).  The features for, e.g., the joint prediction model $g$ of type $\mathcal{CI}$ ($g^{\mathcal{CI}}$) can then simply be described as $\{i \in I | \Phi(i) =\mathcal{CI}\}$. Recall that features computed on the basis of the EAU span are \textit{content-based} ($\mathcal{CB}$), features from the EAU-surrounding text are \textit{content-ignorant} ($\mathcal{CI}$) and features computed from both are denoted by \textit{full-access} ($\mathcal{FA}$). Details on the extraction of features are provided below.

\paragraph{Lexical features}  These consist of boolean values indicating whether a certain word appears in the argumentative source or target EAU or both (and separately, their contexts). More precisely, for any classification instance
we extract uni-grams from within the span of the EAU (if $\mathcal{T}=\mathcal{CB}$)
or solely from the sentence-context surrounding the EAUs (if $\mathcal{T}=\mathcal{CI}$).
Words which occur in both bags are only visible in the full-access setup $\mathcal{T}=\mathcal{FA}$ and are 
modeled as binary indicators.

\begin{figure}
    \centering
    \includegraphics[scale=0.47]{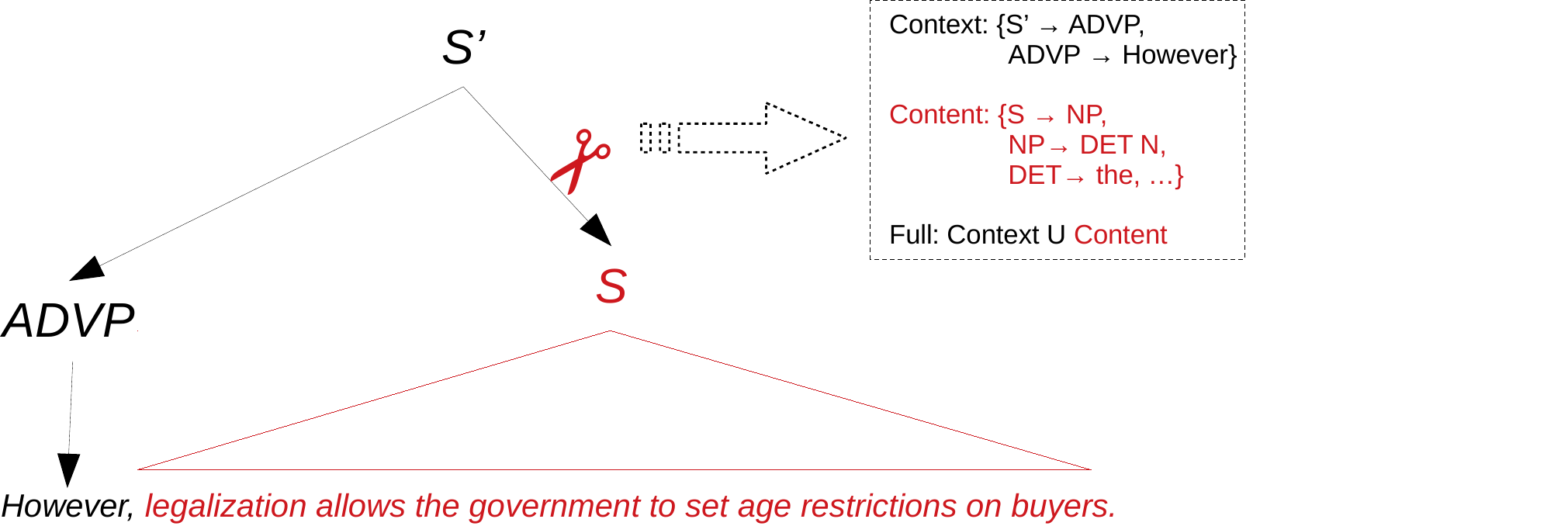}
    \caption{Production rule extraction from constituency parse for two different argumentative units. 
    }
    \label{fig:treefeaextract}
\end{figure}

\paragraph{Syntactic features} 
Such features consist of syntactic production rules extracted from constituency trees -- they are modelled analogously to the lexical features as a bag of production rules. 
To make a clear division between features derived from the EAU embedding context and features derived from within the EAU span, we divide the constituency tree in two parts, as is illustrated in Figure \ref{fig:treefeaextract}. %
If the EAU is embedded in a covering sentence, we cut the syntax tree at the corresponding edge (\includegraphics[scale=0.5]{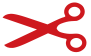} in Figure \ref{fig:treefeaextract}). 
In this
example, the content-ignorant ($\mathcal{CI}$) bag-of-word production rule representation 
includes the rules $S \rightarrow ADVP$ and $ADVP \rightarrow however$. Analogously to the lexical features, the production rules are modeled as binary indicator features.\footnote{A notable insight from our experiments is that the
production rule features have a considerable 
intersection with lexical features. This is due to the terminal production rules, which correspond to
the leaves of the constituency tree.
This explains the surprisingly high scores for production rule features in the production-rule-only ablation experiments in e.g., \citet{stab2014identifying,DBLP:journals/corr/StabG16}.}

\paragraph{Structural} These features describe shallow statistics such as the ratio of argumentative unit tokens compared to sentence tokens or the position of the argumentative unit in the paragraph. We set these features to zero for the content representation of the argumentative unit and replicate those features that allow us to 
treat the argumentative unit as a black-box. For example, in the content-based ($\mathcal{CB}$) system that has access only to the EAU, we can compute the \#tokens in the EAU, but not the  \#tokens in EAU divided by \#tokens in the sentence. The latter feature is only accessible in the full access system variants. 
Hence, in the content-based ($\mathcal{CB}$) system most of these statistics are set to zero since they cannot be computed by considering only the EAU span.

\paragraph{Discourse} For the content-based representation  we retrieve only 
discourse relations that
are confined within the span of the argumentative unit. In the very frequent case that discourse features cross the boundaries of embedding context and EAU span, we only take them into account for $\mathcal{FA}$.

\paragraph{Embeddings} We use the element-wise sum of 300-dimensional pre-trained GloVe vectors \cite{pennington2014glove} corresponding to the words within the EAU span 
($\mathcal{CB}$) and the words of the EAU-surrounding
context 
($\mathcal{CI}$). 
Additionally, we compute the element-wise subtraction of the source EAU vector from the target EAU vector, with the aim of modelling directions in distributional space, similarly to \citet{mikolov2013distributed}. Words with no corresponding pre-trained word vector and empty sequences (e.g., no preceding context available) are treated as a zero-vector.

\paragraph{Sentiment} Tree-based sentiment 
annotations are sentiment scores assigned to nodes 
in 
constituency parse trees \citep{socher-etal-2013-recursive}. We represent these scores
by a one-hot vector of dimension 5 (5 is very positive, 1 is very negative). We determine the \textit{contextual} ($\mathcal{CI}$) sentiment by looking at the highest possible node of the context which does not contain the EAU (ADVP in Figure \ref{fig:treefeaextract}). 
The sentiment for an EAU span ($\mathcal{CB}$) is assigned to the highest possible node covering the EAU span which
does not contain the context sub-tree (S in Figure \ref{fig:treefeaextract}). The full-access ($\mathcal{FA}$) score is assigned to the lowest 
possible node which covers both the EAU span and its surrounding context (S' in Figure \ref{fig:treefeaextract}). Next to the sentiment scores for the selected tree nodes and 
analogously to the word embeddings, we also calculate the element-wise subtraction of the one-hot sentiment source vectors from the one-hot sentiment target vectors. This results in
three additional vectors corresponding to $\mathcal{CB}$, $\mathcal{CI}$ and $\mathcal{FA}$ difference vectors. %

\section{Experiments}
\label{sec:experiments}
\paragraph{Data and pre-processing} We use the corpus of 402 persuasive essays which were annotated with argumentative units, their stances towards the topic and argumentative relations \citep{DBLP:journals/corr/StabG16}. The data is suited for our experiments because the annotators were explicitly asked to provide  annotations on a clausal level. This entails 
that contextual clues tend not to be contained in the annotated span (e.g., only \textit{people should not smoke} is annotated as EAU in the sentence \textit{Therefore, people should not smoke.}). In this work, we are concerned with classifying relations between argumentative units into \textit{support} or \textit{attack} and thus do not consider other annotations. For feature extraction, we process all documents with Stanford CoreNLP \citep{manning-etal-2014-stanford} with the following annotation layers: \textit{sentence tokenize, word tokenize, constituency parse} and \textit{constituency-sentiment}. For extraction of the discourse-features, we proceed by parsing all documents with the PDTB-parser\footnote{\url{https://github.com/WING-NUS/pdtb-parser}} developed by \citet{DBLP:journals/corr/abs-1011-0835}. For the joint task of predicting three link classes (including a non-linked class), we extract as non-linked EAU pairs all EAU pairs which are not linked on a document level. Data set statistics are displayed in Table \ref{tab:datastats}. 

\begin{table}
    \centering
    \begin{tabular}{l|rr|rr}
         &\multicolumn{2}{c}{\#train}&\multicolumn{2}{c}{\#test}\\
         \midrule
         model &$h$ \& $f$ & $g$&$h$ \& $f$ & $g$\\
         \midrule
        documents & 322 & 322 &80&80 \\
        support  & 3820&3820&1021&1021\\
        attack  & 405&405&92&92\\
       $\varnothing$ & - & 5474& -& 1622\\
    \end{tabular}
    \caption{Data set statistics.}
    \label{tab:datastats}
\end{table}

\paragraph{Setup} As explained in \S \ref{sec:argmodels}, we are interested in three distinct configurations of the argumentative relation classifier: \textbf{content-based ($\mathcal{CB}$)}, \textbf{content-ignorant ($\mathcal{CI}$)} and \textbf{full-access ($\mathcal{FA}$)}. Naturally, we would expect the latter to perform best and perhaps we would also expect $\mathcal{CB}$ to outperform $\mathcal{CI}$ -- a system which has no access to the argumentative unit internals whatsoever should not be able to confidently determine relations between them.
Note that some features are only available to $\mathcal{FA}$, which is the case when features cross context and argumentative unit spans (e.g., some of the discourse features), thereby resisting a clear categorization into $\mathcal{CB}$ or $\mathcal{CI}$. Same as most prior work, we use an SVM to learn the feature weights. 

\subsection{Results} 

\paragraph{Replication experiments} Our first step towards our main experiments is 
to replicate the competitive argumentative relation classifier of \citet{DBLP:journals/corr/StabG16,stab2014identifying}. Hence, for comparison purposes, we first formulate the task exactly as it was done in this prior work, using
the model formulation in Eq.\ \ref{eq:1}, which determines the type of outgoing edge from a source (i.e., tree-like view). 

\begin{table}
    \centering
    \begin{tabular}{lrrr}
    \toprule
        system & F1$_{sup}$ & F1$_{att}$ & macro F1 \\
        \midrule
        S\&G16 & 94.7& 41.3&68.0\\
        replicated ($h^{\mathcal{FA}}$)& 94.7 &44.0 &  69.3\\
        \bottomrule
    \end{tabular}
    \caption{Baseline system replication results. 
    }
    \label{tab:repli}
\end{table}
The 
results in Table \ref{tab:repli} confirm the
results 
of \citet{DBLP:journals/corr/StabG16} and suggest that we 
successfully replicated a large proportion of their features.

\begin{table}
    \centering
    \scalebox{0.93}{
    \begin{tabular}{llllll}
    \toprule
        model & $\mathcal{T}$ & F1$_{sup}$ & F1$_{att}$ & F1$_{\varnothing}$ & macro F1 \\
        \midrule
        $h$ & mfs & 95.7&0& -& 47.8\\
         &$\mathcal{CB}$ & 92.9&21.7& -& 57.3$^{\dagger}$\\
         &$\mathcal{CI}$ & 95.0 &38.6 & -&  67.0$^{\dagger\ddagger}$\\
          &$\mathcal{FA}$  & 94.7 &44.0 & -&  69.3$^{\dagger\ddagger}$\\
          \midrule
          $f$ & mfs & 95.7&0& -& 47.8\\
           &$\mathcal{CB}$ & 92.3&20.3& -& 56.3$^{\dagger}$\\
         &$\mathcal{CI}$ & 96.1 &41.7 & -&  70.8$^{\dagger\ddagger}$\\
          &$\mathcal{FA}$ &94.4 &42.4&-&68.5$^{\dagger\ddagger}$\\
          \midrule
          
          $g$ & mfs & 0&0& 74.5& 8.3\\
          &$\mathcal{CB}$ & 54.3&9.9& 65.0& 43.4$^{\dagger}$\\
         &$\mathcal{CI}$ & 63.0 &34.8 & 76.5 &59.3$^{\dagger\ddagger}$\\
          &$\mathcal{FA}$ &46.6 &32.3 & 73.1 &56.1$^{\dagger\ddagger}$\\
        \bottomrule
    \end{tabular}}
    \caption{
    Argumentative relation classification models $h, f, g$ with different access to content and context; models of type $\mathcal{CI}$ (content-ignorant) have no access to the EAU span. $\dagger$: significantly better than mfs baseline ($p<0.005$);  $\ddagger$ significantly better than content-based ($p<0.005$). 
    }
    \label{tab:mainres}
\end{table}

\paragraph{Main results} The 
results for all three prediction settings (one outgoing edge: $h$, support/attack: $f$ and support/attack/neither: $g$) across all type variables 
($\mathcal{CB}$, $\mathcal{CI}$ and $\mathcal{FA}$) are displayed in Table \ref{tab:mainres}. 
All models significantly outperform the majority baseline with respect to macro F1. Intriguingly, the content-ignorant models ($\mathcal{CI}$) \textit{always} perform significantly better than the models which only have access to the EAUs'
content ($\mathcal{CB}$, $p<0.005$). In the most general task formulation ($g$), we 
observe that $\mathcal{CI}$ even significantly outperforms the model which has maximum access (seeing both EAU spans and surrounding 
contexts: $\mathcal{FA}$).

 At first glance, the results of the purely EAU focused systems ($\mathcal{CB}$) are disappointing, 
since they fall far behind their
 competitors. On the other hand, their
 F1 scores are not devastatingly bad. The strong most-frequent-class baseline 
 is significantly outperformed by the content-based ($\mathcal{CB}$) system, across all three prediction settings.

In summary
our findings are as follows: (i) models which see the EAU span (content-based, $\mathcal{CB}$) are significantly outperformed by models that have no access to the span itself (content-ignorant, $\mathcal{CI}$) across all settings; (ii) in two of three prediction settings ($f$ and $g$), the model which only has access to the context even outperforms the model that has access to all information in the input. 
The fact that using features derived exclusively from the EAU embedding context ($\mathcal{CI}$) can lead to better results than using a full feature-system ($\mathcal{FA}$) suggests that some information from the EAU can even be harmful. Why this is the case, we cannot answer exactly. A plausible cause might be related to the smaller dimension of the feature space, which makes the SVM less likely to overfit. Still, this finding comes as a surprise 
and calls for 
further 
investigation in future work.

\paragraph{Robustness tests} %
A system for argumentative relation classification can be applied in one of two settings: \textit{single-document} or \textit{cross-document}, as illustrated in
Figure \ref{fig:svsm}:
\begin{figure}
    \centering
    \includegraphics[scale=0.50]{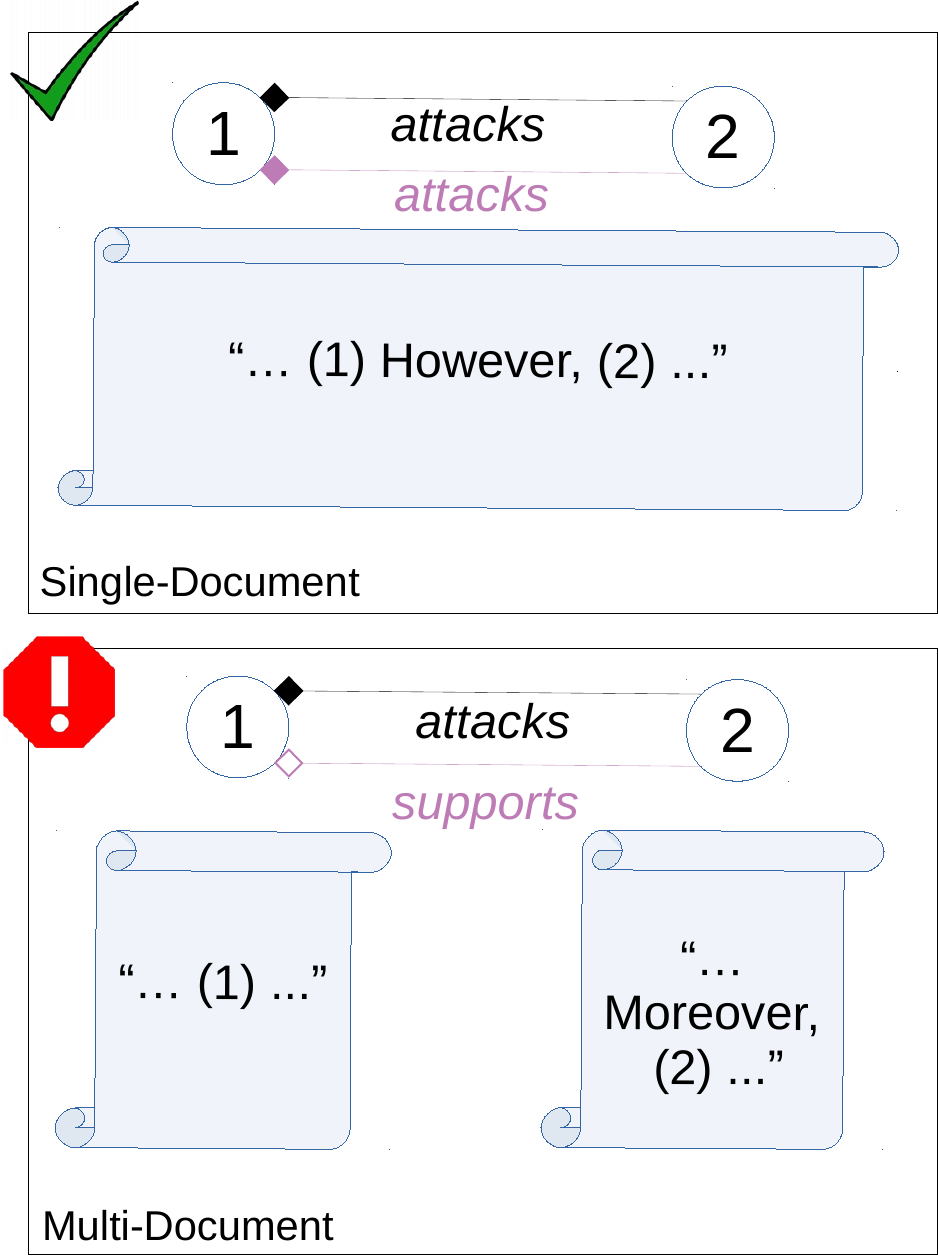}
    \caption{Single-document (top) vs.\ cross-document (bottom) 
    argumentative relation classification. 
    Black edge: gold label; purple edge: predicted label.}
    \label{fig:svsm}
\end{figure}
in the first case (top),
a system is tasked to classify
EAUs that appear linearly in one document -- here 
contextual 
clues can often
highlight
the relationship between two units. This is the setting we have been considering up to now.
However, in the second 
scenario (bottom),
we have moved away from the closed single-document setting and 
ask the system to classify
two EAUs extracted from different document contexts. This setting applies, for instance, when we are mining arguments from multiple sources.

In \textit{both} cases, however, a system that relies more on contextual clues than on the content expressed in the EAUs 
is problematic:
in the single-document setting, such a system will rely on discourse indicators -- whether or not they are justified by content -- and can thus easily be fooled. 


In the cross-document setting,
discourse-based indicators
-- being inherently defined with respect to their internal document context -- do not have a defined rhetorical function with respect to EAUs in a separate document 
and thus a system that
has learned to rely
on 
such markers
within a single-document setting can be seriously misled. 
\begin{figure*}
  \begin{subfigure}{.33\textwidth}
  \centering
    \includegraphics[width=54mm]{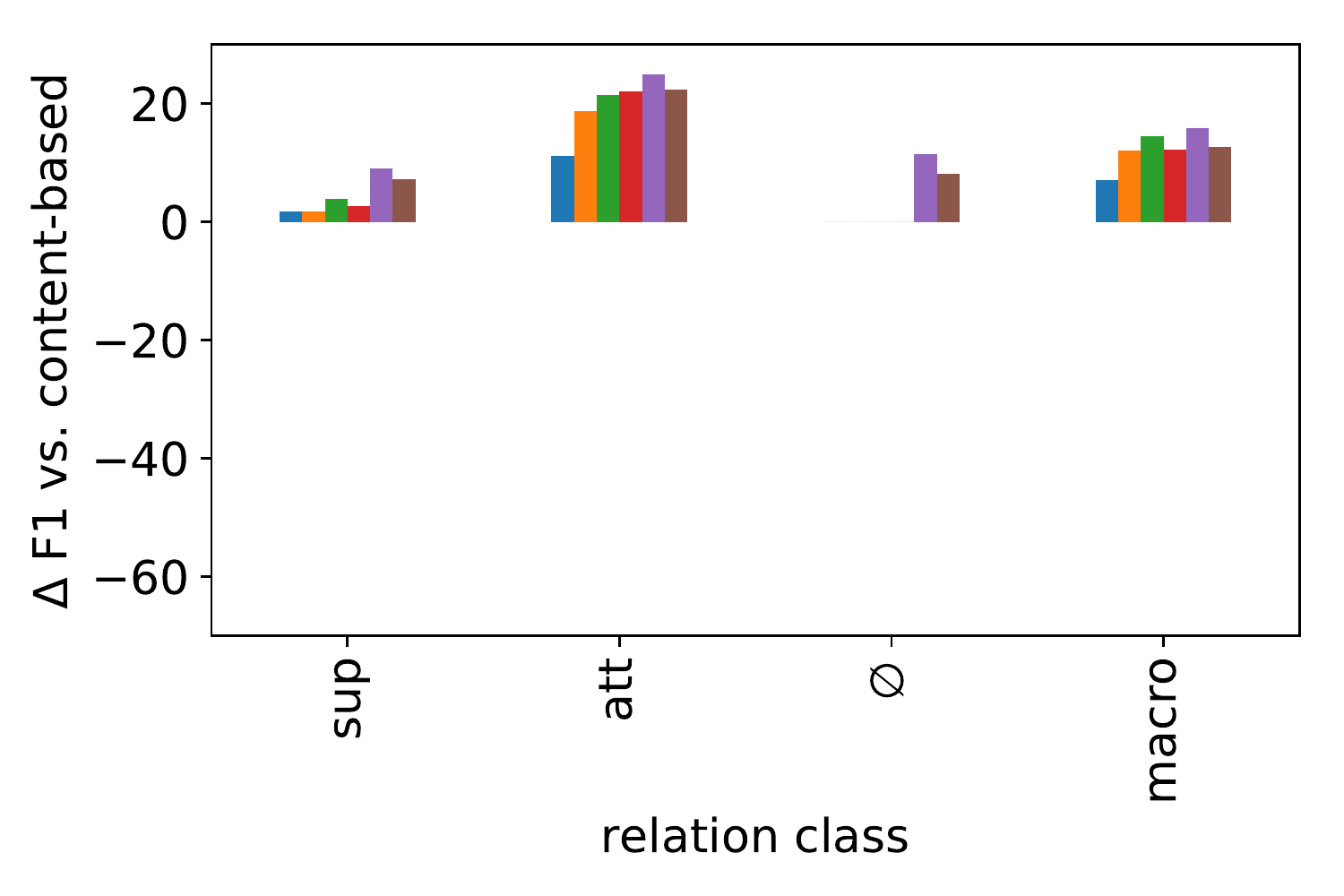}
    \caption{Standard setting.}
    \label{fig:variedtesta}
  \end{subfigure}%
  \begin{subfigure}{.33\textwidth}
  \centering
    \includegraphics[width=54mm]{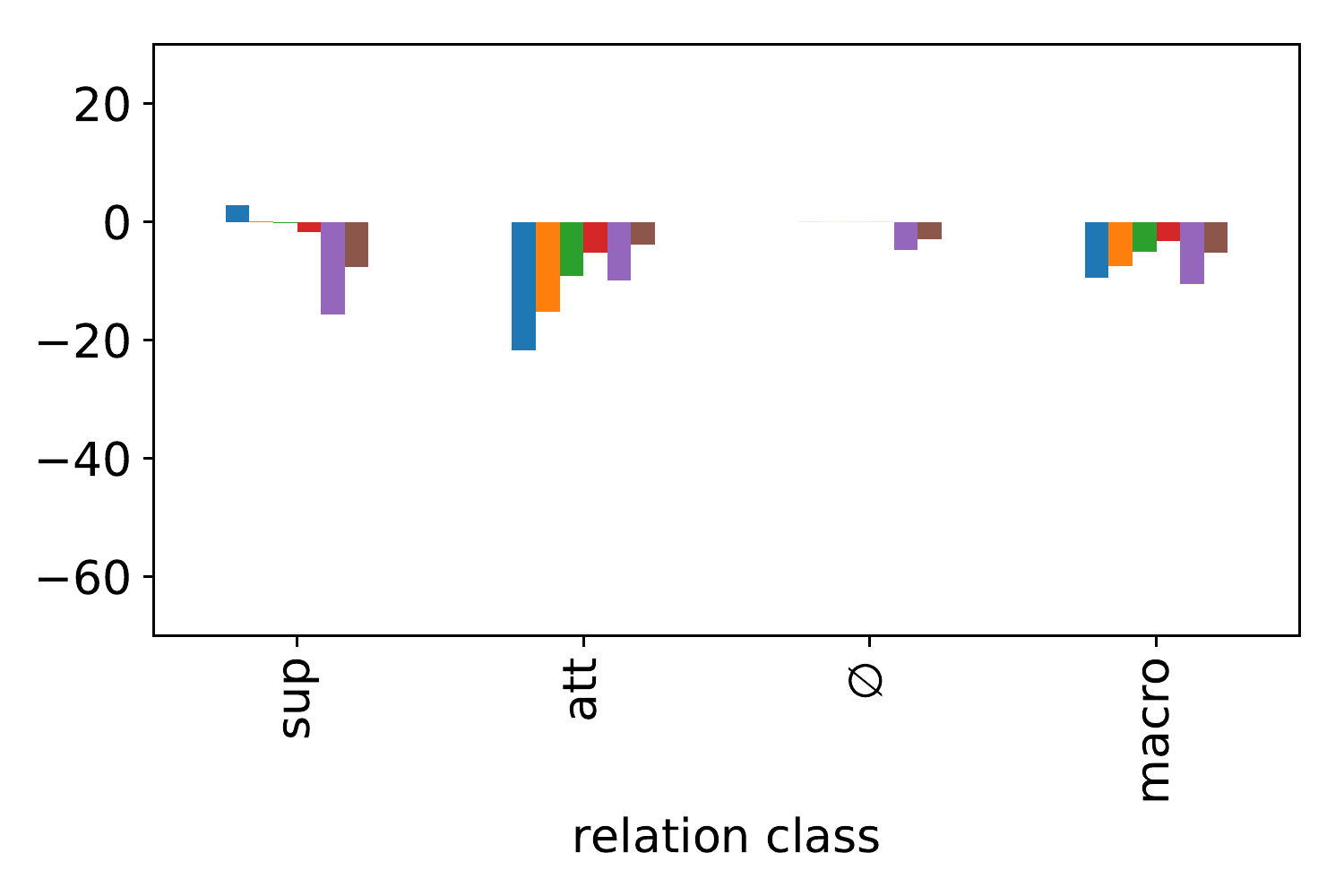}
    \caption{Randomized-context test.}
    \label{fig:variedtestb}
  \end{subfigure}
  \begin{subfigure}{.33\textwidth}
  \centering
    \includegraphics[width=55mm]{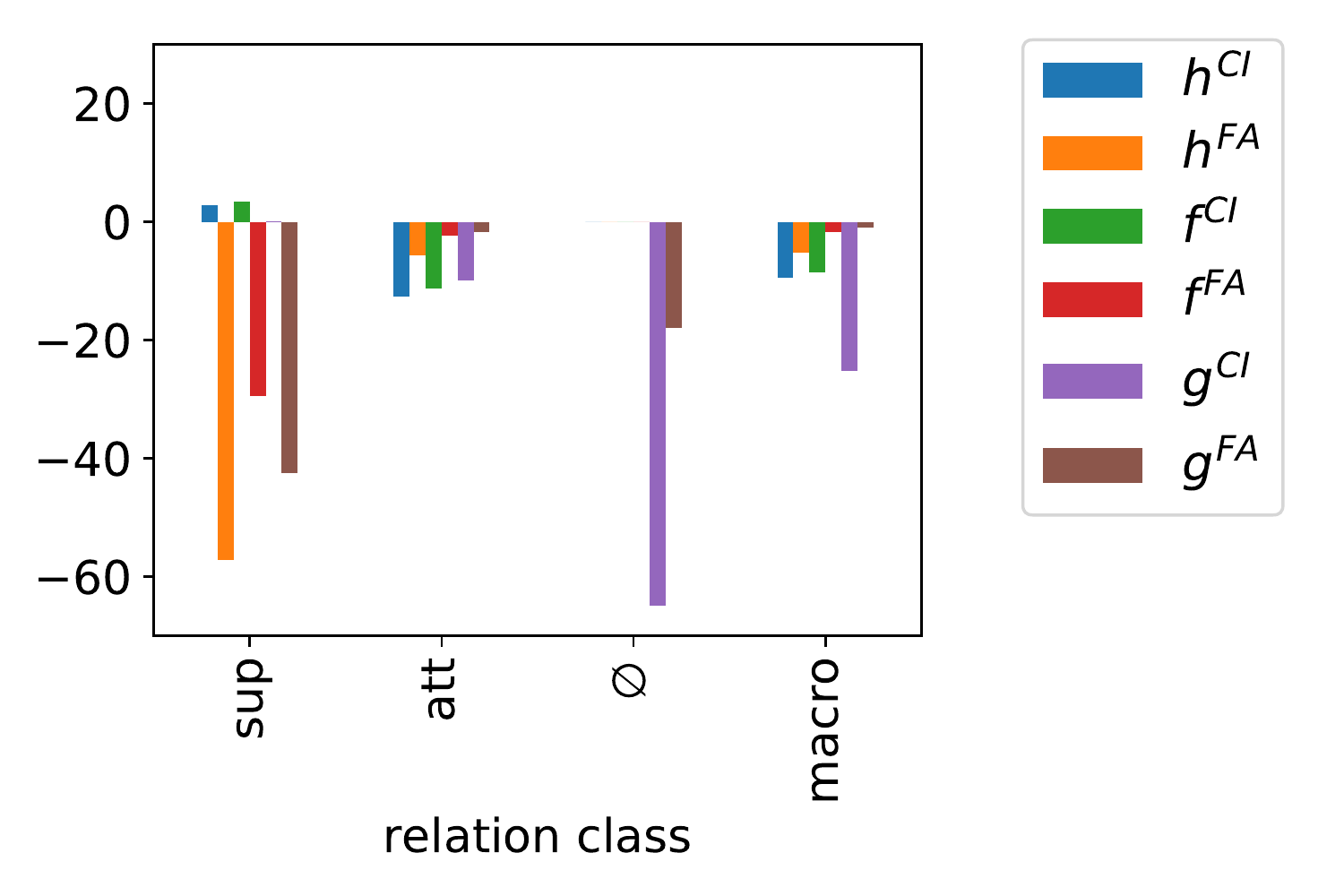}
    \caption{No-context test.}
    \label{fig:variedtestc}
  \end{subfigure}
  
  \caption{\textit{Randomized-context} test set: models are applied to 
 testing instances with randomly flipped contexts.
  \textit{No-context test set}: 
  models can only access the EAU span of a testing instance. 
  A bar below/above zero means that a system that can access context (content-ignorant $\mathcal{CI}$ or full-access $\mathcal{FA}$) is worse/better than the content-based baseline $\mathcal{CB}$ that only has access to the EAU span (its performance is not affected 
  by modified context, cf. Tab.\ \ref{tab:mainres}).}
  \label{fig:variedtest}
 \end{figure*}
We believe that the cross-document
setting 
should be an important goal in argumentation analysis, since 
it generalizes better to many debates of interest, where EAUs can be found scattered across thousands of documents. 
For example, 
for the 
topic of \textit{legalizing marijuana},
EAUs may 
be mined from millions of 
documents and thus their relations may naturally extend across document boundaries. 
If a system learns
to over-proportionally attend to the EAUs'
surrounding contexts it is
prone to making many errors.\footnote{
In fact, similar considerations also apply when moving from argumentative monologue to dialogue, i.e., in interactive debates. Again, systems need to be able to detect relations between EAUs uttered by different speakers and independently from the speaker-specific utterance context.}
%

In what follows we are  simulating the effects that an overly context-sensitive  classifier could have in a cross-document setting, by modifying our experimental setting, and study the effects on the different model types: In one setup -- we call it \textit{randomized-context} -- we systematically distort the context of our testing instances by exchanging the context in a randomized manner; in the other setting -- called \textit{no-context}, we are deleting the context around the ADUs to be classified. \textit{Randomized-context} simulates an open world debate where argumentative units may occur 
in different contexts, sometimes with discourse markers indicating an 
opposite class. In other words, in this setting we want to examine effects when porting a context-sensitive system to a multi-document setting.\footnote{We concede that this is an artificial setup, but defer more realistic cross-document experiments to future work.}
For example, as seen in Figure \ref{fig:svsm}, the context of an argumentative unit may change from ``However'' to ``Moreover'' -- which can happen naturally in open debates. 

The results are displayed in Figure \ref{fig:variedtest}. In the standard setting (Figure \ref{fig:variedtesta}), the models that have 
access to 
the context besides the content ($\mathcal{FA}$) and the models that are only allowed to access the context ($\mathcal{CI}$), always perform better than the content-based models ($\mathcal{CB}$) (bars above zero). However, when we randomly flip
contexts of the test instances (Figure \ref{fig:variedtestb}), or suppress them entirely
(Figure \ref{fig:variedtestc}), the opposite picture emerges: 
the content-based models always 
outperform the other models. For some classes (\textit{support}, $\varnothing$) the difference can 
exceed
50 F1 percentage points. These two studies, where testing examples are 
varied regarding their
context 
(\textit{randomized-context} or \textit{no-context}) simulates what can be expected if we 
apply our systems for relation class assignment 
to
EAUs stemming from heterogeneous sources. While the performances of a purely content-based model naturally stays stable, the performance of the other systems decrease notably -- they perform worse than the content-based model. %
\paragraph{Feature investigation}
\begin{figure}
    \centering
    \includegraphics[width=\linewidth]{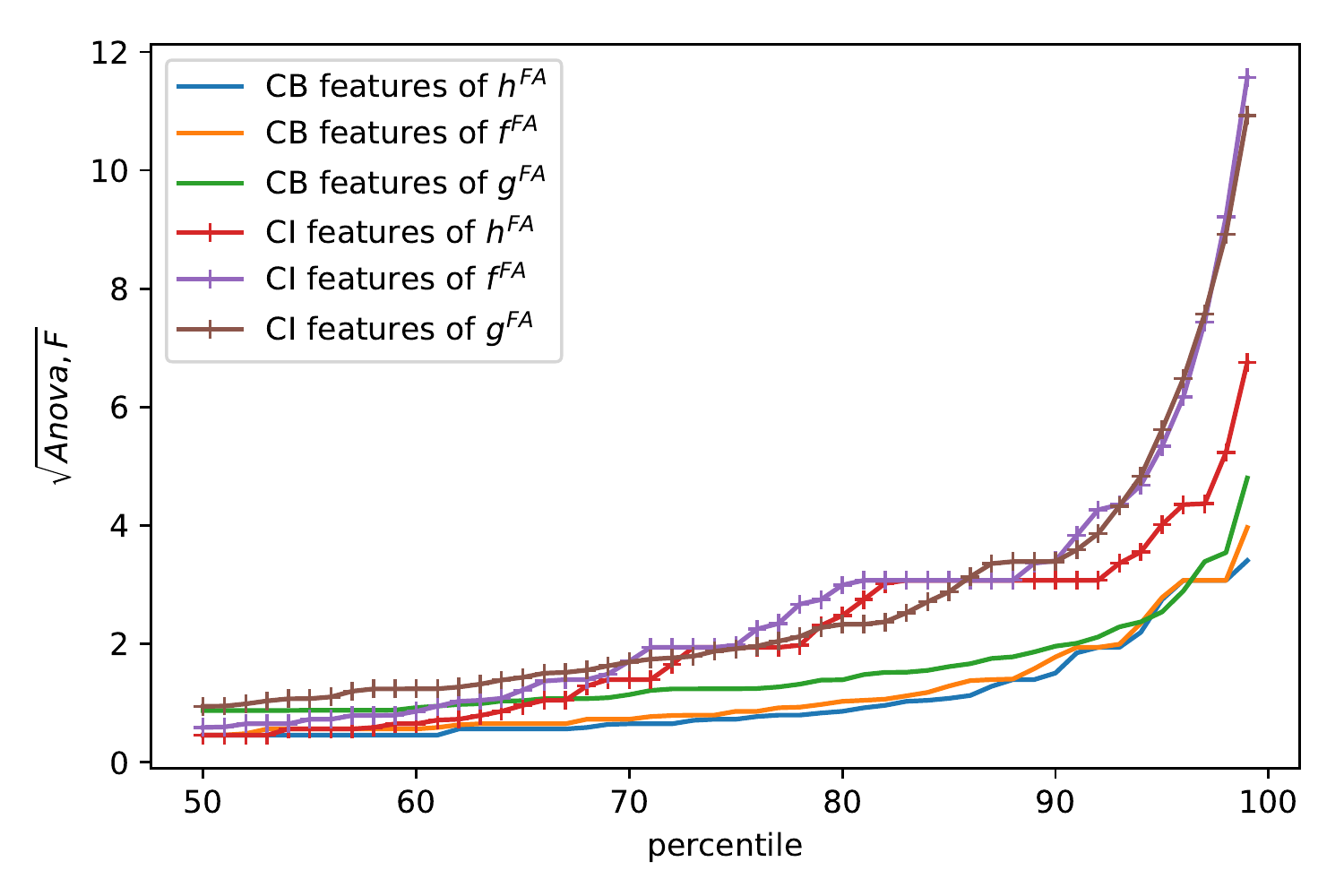}
    \caption{ANOVA F score percentiles for content-based vs.\ content-ignorant features in the training data. A higher feature score suggests greater predictive capacity.}
    \label{fig:anova}
\end{figure}
We calculate the ANOVA classification F scores of the features with respect to our three task formulations $h,g$ and $f$. The F percentiles of features extracted from the EAU surrounding text ($\mathcal{CI}$) and features extracted from the EAU span ($\mathcal{CB}$), are displayed in Figure \ref{fig:anova}.

It clearly stands out that 
features obtained 
from the EAU surrounding context ($\mathcal{CI}$) are assigned much higher scores compared to 
features stemming 
from the EAU span ($\mathcal{CB}$). This 
holds true for all three task formulations and provides further evidence that models -- when given the option -- put a strong focus 
on
contextual clues while neglecting the information provided by the EAU span
itself. 

\section{Discussion}
\label{sec:discussion}

While competitive systems for argumentative relation classification are considered to be robust, our experiments have shown that despite confidence-inspiring scores on unseen testing data, such systems can easily be fooled -- they can deliver strong performance scores although the classifier does not have access to the content of the EAUs. In this
respect, we have provided evidence that there is a
danger in case models focus too much on rhetorical indicators, in detriment of the context. 
Thus, the following 
question 
arises: \textit{How can we prevent argumentation models from modeling arguments or argumentative units and their relations in overly na\"ive ways?} A simple and intuitive way is to dissect 
EAUs from their surrounding document context. Models trained on data that is restricted to the EAUs' content will be forced to 
focus on the content of EAUs.
We believe that this will enhance the robustness of such models and allows them to 
generalize to cross-document argument relation classification.
The corpus of student essays
makes such transformations straightforward: only the EAUs
were annotated (e.g., ``However, $[_{arg}$A$]$''). If annotations extend over the EAUs
(e.g., only full sentences are annotated, ``$[_{arg}$However, A$]$''), 
such transformations 
could be performed 
automatically after a discourse parsing step. When inspecting the student essays corpus, we further observed that an EAU 
mining step should involve coreference resolution to
better capture relations between EAUs
that involve anaphors (e.g., \textit{``Exercising makes you feel better''} and 
\textit{``It$_{[Exercising]}$ increases endorphin levels''}).  

Thus, in order to conduct real-world end-to-end argumentation relation mining for a given topic, we envision a system that addresses 
three
steps: (i) mining of EAUs and (ii) replacement of 
pronouns in EAUs with referenced entities (e.g., $It~is~healthy\rightarrow Excercise~is~healthy$). Finally (iii), given the cross product of mined EAUs we can apply a model of type $g$ to construct a full-fledged argumentation graph, possibly spanning multiple documents.\footnote{cf.\ \citet{peldszus2015joint} for graph prediction within single
documents.} 
We have shown that in order to properly perform step (iii), we need stronger models that are able to 
better model 
EAU contents. Hence, we encourage 
the argumentation community to test their systems on a decontextualized version of the student essays, including the proposed -- and possibly further extended -- testing setups, to challenge the semantic representation and reasoning capacities of argument analysis models.  This will lead to more realistic performance estimates and increased robustness of systems when addressing desirable multi-document tasks.

\section{Conclusion}

We have shown that systems which put too much focus on 
discourse information
may be easily fooled -- an issue which has severe implications when systems are applied to cross-document argumentative relation classification tasks. The strong reliance on contextual clues is also problematic in single-document contexts, where systems can run a risk of assigning relation labels relying on contextual and rhetorical effects -- instead of focusing on content. Hence, we propose that researchers test their argumentative relation classification systems on two alternative versions of the StudentEssay data that reflect different access levels. (i) \textit{EAU-span only}, where systems 
only see the EAU spans and (ii) \textit{context-only}, where
systems can only see the EAU-surrounding context.  These complementary settings will (i) challenge the semantic capacities of a system, and (ii) unveil the extent to which a system is focusing on the discourse context when making decisions. We will offer our 
testing environments to the research community through
a platform that provides  datasets and scripts and 
a table to trace the results of content-based systems.\footnote{\url{http://explain.cl.uni-heidelberg.de/}}

\section*{Acknowledgments}

This work has been supported by the German Research Foundation as part of the Research Training Group Adaptive Preparation of Information from Heterogeneous  Sources  (AIPHES)  under  grant no.\ GRK 1994/1 and by  the  Leibniz  ScienceCampus ``Empirical  Linguistics  and  Computational  Language  Modeling'', supported by the Leibniz Association under grant no.\  SAS-2015-IDS-LWC  and  by  the  Ministry  of  Science, Research, and Art of Baden-W\"urttemberg.
\bibliography{acl2019}

\begin{thebibliography}{29}
\expandafter\ifx\csname natexlab\endcsname\relax\def\natexlab#1{#1}\fi

\bibitem[{Aker et~al.(2017)Aker, Sliwa, Ma, Lui, Borad, Ziyaei, and
  Ghobadi}]{aker-etal-2017-works}
Ahmet Aker, Alfred Sliwa, Yuan Ma, Ruishen Lui, Niravkumar Borad, Seyedeh
  Ziyaei, and Mina Ghobadi. 2017.
\newblock \href {https://doi.org/10.18653/v1/W17-5112} {What works and what
  does not: Classifier and feature analysis for argument mining}.
\newblock In \emph{Proceedings of the 4th Workshop on Argument Mining}, pages
  91--96, Copenhagen, Denmark.

\bibitem[{Azar(1999)}]{Azar1999}
M.~Azar. 1999.
\newblock \href {https://doi.org/10.1023/A:1007794409860} {Argumentative text
  as rhetorical structure: An application of rhetorical structure theory}.
\newblock \emph{Argumentation}, 13(1):97--114.

\bibitem[{Cocarascu and Toni(2017)}]{cocarascu2017identifying}
Oana Cocarascu and Francesca Toni. 2017.
\newblock Identifying attack and support argumentative relations using deep
  learning.
\newblock In \emph{Proceedings of the 2017 Conference on Empirical Methods in
  Natural Language Processing}, pages 1374--1379.

\bibitem[{Green et~al.(2014)Green, Ashley, Litman, Reed, and
  Walker}]{W14-21:2014}
Nancy Green, Kevin Ashley, Diane Litman, Chris Reed, and Vern Walker, editors.
  2014.
\newblock \href {http://www.aclweb.org/anthology/W/W14/W14-21}
  {\emph{Proceedings of the First Workshop on Argumentation Mining}}.
\newblock Association for Computational Linguistics, Baltimore, Maryland.

\bibitem[{Green(2010)}]{green2010representation}
Nancy~L Green. 2010.
\newblock Representation of argumentation in text with rhetorical structure
  theory.
\newblock \emph{Argumentation}, 24(2):181--196.

\bibitem[{Hou and Jochim(2017)}]{hou2017argument}
Yufang Hou and Charles Jochim. 2017.
\newblock Argument relation classification using a joint inference model.
\newblock In \emph{Proceedings of the 4th Workshop on Argument Mining}, pages
  60--66.

\bibitem[{Kuribayashi et~al.(2018)Kuribayashi, Reisert, Inoue, and
  Inui}]{kuri-2018}
Tatsuki Kuribayashi, Paul Reisert, Naoya Inoue, and Kentaro Inui. 2018.
\newblock Towards exploiting argumentative context for argumentative relation
  identification.
\newblock In \emph{Proceedings of the Annual Meeting of the Association for
  Natural Language Processing NLP}, pages 284--287.

\bibitem[{Levy et~al.(2014)Levy, Bilu, Hershcovich, Aharoni, and
  Slonim}]{levy-etal-2014-context}
Ran Levy, Yonatan Bilu, Daniel Hershcovich, Ehud Aharoni, and Noam Slonim.
  2014.
\newblock \href {https://www.aclweb.org/anthology/C14-1141} {Context dependent
  claim detection}.
\newblock In \emph{Proceedings of {COLING} 2014, the 25th International
  Conference on Computational Linguistics: Technical Papers}, pages 1489--1500,
  Dublin, Ireland. Dublin City University and Association for Computational
  Linguistics.

\bibitem[{Lin et~al.(2014)Lin, Ng, and Kan}]{DBLP:journals/corr/abs-1011-0835}
Ziheng Lin, Hwee~Tou Ng, and Min-Yen Kan. 2014.
\newblock A pdtb-styled end-to-end discourse parser.
\newblock \emph{Natural Language Engineering}, 20(2):151--184.

\bibitem[{Lippi and Torroni(2016)}]{lippi2016argumentation}
Marco Lippi and Paolo Torroni. 2016.
\newblock Argumentation mining: State of the art and emerging trends.
\newblock \emph{ACM Transactions on Internet Technology (TOIT)}, 16(2):10.

\bibitem[{Mann and Thompson(1987)}]{mann1987rhetorical}
William~C Mann and Sandra~A Thompson. 1987.
\newblock Rhetorical structure theory: Description and construction of text
  structures.
\newblock In \emph{Natural language generation}, pages 85--95. Springer.

\bibitem[{Manning et~al.(2014)Manning, Surdeanu, Bauer, Finkel, Bethard, and
  McClosky}]{manning-etal-2014-stanford}
Christopher Manning, Mihai Surdeanu, John Bauer, Jenny Finkel, Steven Bethard,
  and David McClosky. 2014.
\newblock \href {https://doi.org/10.3115/v1/P14-5010} {The {S}tanford
  {C}ore{NLP} natural language processing toolkit}.
\newblock In \emph{Proceedings of 52nd Annual Meeting of the Association for
  Computational Linguistics: System Demonstrations}, pages 55--60, Baltimore,
  Maryland.

\bibitem[{Mikolov et~al.(2013)Mikolov, Sutskever, Chen, Corrado, and
  Dean}]{mikolov2013distributed}
Tomas Mikolov, Ilya Sutskever, Kai Chen, Greg~S Corrado, and Jeff Dean. 2013.
\newblock Distributed representations of words and phrases and their
  compositionality.
\newblock In \emph{Advances in neural information processing systems}, pages
  3111--3119.

\bibitem[{Mochales and Moens(2011)}]{mochales2011argumentation}
Raquel Mochales and Marie-Francine Moens. 2011.
\newblock Argumentation mining.
\newblock \emph{Artificial Intelligence and Law}, 19(1):1--22.

\bibitem[{Nguyen(2018)}]{nguyen2018context}
Huy Nguyen. 2018.
\newblock \emph{Context-aware Argument Mining and Its Applications in
  Education}.
\newblock Ph.D. thesis, University of Pittsburgh.

\bibitem[{Nguyen and Litman(2016)}]{nguyen2016context}
Huy Nguyen and Diane Litman. 2016.
\newblock Context-aware argumentative relation mining.
\newblock In \emph{Proceedings of the 54th Annual Meeting of the Association
  for Computational Linguistics (Volume 1: Long Papers)}, volume~1, pages
  1127--1137.

\bibitem[{Peldszus and Stede(2013)}]{peldszus2013argument}
Andreas Peldszus and Manfred Stede. 2013.
\newblock From argument diagrams to argumentation mining in texts: A survey.
\newblock \emph{International Journal of Cognitive Informatics and Natural
  Intelligence (IJCINI)}, 7(1):1--31.

\bibitem[{Peldszus and Stede(2015)}]{peldszus2015joint}
Andreas Peldszus and Manfred Stede. 2015.
\newblock Joint prediction in mst-style discourse parsing for argumentation
  mining.
\newblock In \emph{Proceedings of the 2015 Conference on Empirical Methods in
  Natural Language Processing}, pages 938--948.

\bibitem[{Peldszus and Stede(2016)}]{PeldszusStede-ECA:16}
Andreas Peldszus and Manfred Stede. 2016.
\newblock An annotated corpus of argumentative microtexts.
\newblock In D.~Mohammed and M.~Lewinski, editors, \emph{Argumentation and
  Reasoned Action - Proc. of the 1st European Conference on Argumentation,
  Lisbon, 2015}. College Publications, London.

\bibitem[{Pennington et~al.(2014)Pennington, Socher, and
  Manning}]{pennington2014glove}
Jeffrey Pennington, Richard Socher, and Christopher Manning. 2014.
\newblock Glove: Global vectors for word representation.
\newblock In \emph{Proceedings of the 2014 conference on Empirical Methods in
  Natural Language Processing (EMNLP)}, pages 1532--1543.

\bibitem[{Persing and Ng(2016)}]{persing-ng-2016-end}
Isaac Persing and Vincent Ng. 2016.
\newblock \href {https://doi.org/10.18653/v1/N16-1164} {End-to-end
  argumentation mining in student essays}.
\newblock In \emph{Proceedings of the 2016 Conference of the North {A}merican
  Chapter of the Association for Computational Linguistics: Human Language
  Technologies}, pages 1384--1394, San Diego, California. Association for
  Computational Linguistics.

\bibitem[{Rasooli and Tetreault(2015)}]{rasooli-tetrault-2015}
Mohammad~Sadegh Rasooli and Joel~R. Tetreault. 2015.
\newblock \href {http://arxiv.org/abs/1503.06733} {Yara parser: {A} fast and
  accurate dependency parser}.
\newblock \emph{Computing Research Repository}, arXiv:1503.06733.
\newblock Version 2.

\bibitem[{Richardson and Domingos(2006)}]{richardson2006markov}
Matthew Richardson and Pedro Domingos. 2006.
\newblock Markov logic networks.
\newblock \emph{Machine learning}, 62(1-2):107--136.

\bibitem[{Socher et~al.(2013)Socher, Perelygin, Wu, Chuang, Manning, Ng, and
  Potts}]{socher-etal-2013-recursive}
Richard Socher, Alex Perelygin, Jean Wu, Jason Chuang, Christopher~D. Manning,
  Andrew Ng, and Christopher Potts. 2013.
\newblock \href {https://www.aclweb.org/anthology/D13-1170} {Recursive deep
  models for semantic compositionality over a sentiment treebank}.
\newblock In \emph{Proceedings of the 2013 Conference on Empirical Methods in
  Natural Language Processing}, pages 1631--1642, Seattle, Washington, USA.

\bibitem[{Stab and
  Gurevych(2014{\natexlab{a}})}]{stab-gurevych-2014-annotating}
Christian Stab and Iryna Gurevych. 2014{\natexlab{a}}.
\newblock \href {https://www.aclweb.org/anthology/C14-1142} {Annotating
  argument components and relations in persuasive essays}.
\newblock In \emph{Proceedings of {COLING} 2014, the 25th International
  Conference on Computational Linguistics: Technical Papers}, pages 1501--1510,
  Dublin, Ireland. Dublin City University and Association for Computational
  Linguistics.

\bibitem[{Stab and Gurevych(2014{\natexlab{b}})}]{stab2014identifying}
Christian Stab and Iryna Gurevych. 2014{\natexlab{b}}.
\newblock Identifying argumentative discourse structures in persuasive essays.
\newblock In \emph{Proceedings of the 2014 Conference on Empirical Methods in
  Natural Language Processing (EMNLP)}, pages 46--56.

\bibitem[{Stab and Gurevych(2017)}]{DBLP:journals/corr/StabG16}
Christian Stab and Iryna Gurevych. 2017.
\newblock \href {https://doi.org/10.1162/COLI_a_00295} {Parsing argumentation
  structures in persuasive essays}.
\newblock \emph{Computational Linguistics}, 43(3):619--659.

\bibitem[{Stab et~al.(2018)Stab, Miller, Schiller, Rai, and
  Gurevych}]{stab2018cross}
Christian Stab, Tristan Miller, Benjamin Schiller, Pranav Rai, and Iryna
  Gurevych. 2018.
\newblock Cross-topic argument mining from heterogeneous sources.
\newblock In \emph{Proceedings of the 2018 Conference on Empirical Methods in
  Natural Language Processing}, pages 3664--3674.

\bibitem[{Wachsmuth et~al.(2018)Wachsmuth, Stede, El~Baff, Al~Khatib,
  Skeppstedt, and Stein}]{wachsmuth2018argumentation}
Henning Wachsmuth, Manfred Stede, Roxanne El~Baff, Khalid Al~Khatib, Maria
  Skeppstedt, and Benno Stein. 2018.
\newblock Argumentation synthesis following rhetorical strategies.
\newblock In \emph{Proceedings of the 27th International Conference on
  Computational Linguistics}, pages 3753--3765.

\end{thebibliography}
\bibliographystyle{acl_natbib}

\end{document}